\pdfoutput=1

\documentclass[11pt]{article}

\usepackage{graphicx}

\usepackage{color,soul}
\definecolor{orange}{rgb}{1,0.45,0}
\setulcolor{orange}

\usepackage{comment}

\definecolor{eventA}{rgb}{0.13,0.44,0.75}
\setulcolor{eventA}

\usepackage[]{2023}

\usepackage{times}
\usepackage{latexsym}

\usepackage[T1]{fontenc}

\usepackage[utf8]{inputenc}

\usepackage{microtype}

\usepackage{inconsolata}

\newcommand{\myparagraph}[1]{\vspace{2pt}\noindent{\textbf{#1}}}

\newcommand\blfootnote[1]{%
  \begingroup
  \renewcommand\thefootnote{}\footnote{#1}%
  \addtocounter{footnote}{-1}%
  \endgroup
}

\title{Do large language models and humans have similar behaviors in causal inference with script knowledge?}

\author{
  Xudong Hong\footnotemark[1] $^{1,2}$, Margarita Ryzhova\footnotemark[1] $^{2}$, Daniel Adrian Biondi$^{2}$ and Vera Demberg$^{1,2}$ \\[0.8ex]
  $^{1}$Dept. of Computer Science, Saarland University \\
  $^{2}$Dept. of Language Science and Technology, Saarland University \\
  {\tt \{xhong,mryzhova,biondi,vera\}@lst.uni-saarland.de}\\
}

\begin{document}
\maketitle
\blfootnote{$\ast$ These authors contributed equally to this work.}

\begin{abstract}
Recently, large pre-trained language models (LLMs) have demonstrated superior language understanding abilities, including zero-shot causal reasoning. However, it is unclear to what extent their capabilities are similar to human ones. We here study the processing of an event $B$ in a script-based story, which causally depends on a previous event $A$. In our manipulation, event $A$ is stated, negated, or omitted in an earlier section of the text. 
We first conducted a self-paced reading experiment, which showed that humans exhibit significantly longer reading times when causal conflicts exist ($\neg A \rightarrow B$) than under logical conditions ($A \rightarrow B$). However, reading times remain similar when cause A is not explicitly mentioned, indicating that humans can easily infer event B from their script knowledge. 
We then tested a variety of LLMs on the same data to check to what extent the models replicate human behavior. 
Our experiments show that 1) only recent LLMs, like GPT-3 or Vicuna, correlate with human behavior in the $\neg A \rightarrow B$ condition. 
2) Despite this correlation, all models still fail to predict that  $nil \rightarrow B$ is less surprising than $\neg A \rightarrow B$, indicating that LLMs still have difficulties integrating script knowledge. Our code and collected data set are available at \url{https://github.com/tony-hong/causal-script}.
\end{abstract}

\section{Introduction}
Causal reasoning is fundamental for both human and machine intelligence \citep{pearl2009causality} and plays an important role in language comprehension \citep{keenan1974identification,graesser1994constructing, graesser1997discourse,van1990causal}. Large pre-trained language models (LLMs) such as GPT-3.5 \citep{neelakantan2022gpt3.5} have demonstrated excellent zero-shot capabilities in causal reasoning tasks and human-like behaviors \citep{wang2019superglue,kojima2022zeroshotcot}. 
On the other hand, some early pieces of evidence show that LLMs lack global planning of different events in long texts \citep{bubeck2023sparks}. So it is unclear to what extent LLMs can conduct causal reasoning about events. 

\begin{figure*}[ht]
    \centering
    \includegraphics[width=\textwidth, clip=TRUE, trim=0cm 0cm 2.02cm 0cm]{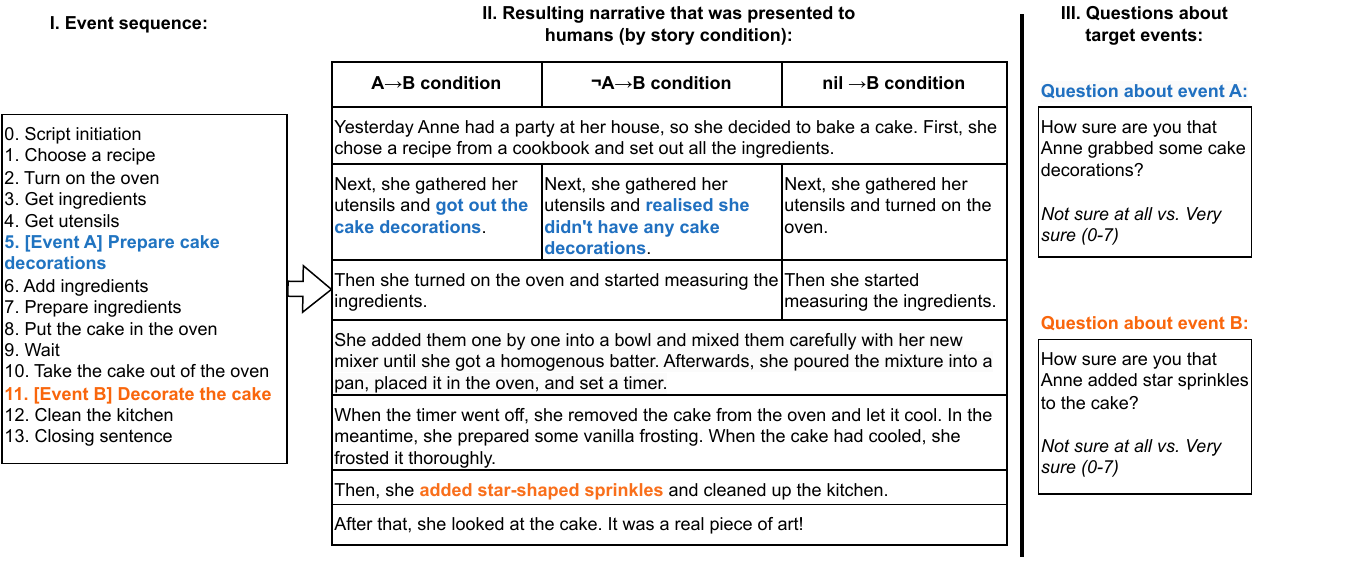}
    \caption{Example of a script structure (I), the resulting narrative in three conditions (II) and questions that subjects were asked (III), for "baking a cake" story. }
    \label{fig:example_materials}
\end{figure*}

In turn, humans have been shown to be extremely good at building causal connections in long discourse comprehension \citep{radvansky2014different, graesser1994constructing}. In doing so, they rely not only on explicit causal links \citep[signalled in the text -- see ][]{trabasso1985causal, keenan1974identification} but also on implicit ones that are inferable based on commonsense knowledge \citep{keenan1974identification, singer1996constructing}. In particular, subjects were found to be sensitive to causal conflicts arising when something in the text contradicts either what was written before or subjects' commonsense knowledge \citep{radvansky2014different, singer1996constructing}. An example of a causal conflict is presented in Figure \ref{fig:example_materials}, Part II, condition $\neg A \rightarrow B$, where decorating a cake with star-shaped sprinkles is inconsistent with the previously mentioned information that cake decorations are not available. 

In this paper, we investigate language processing in humans and compare it to a large variety of LLMs, following the ``psycholinguistic assessment of language models paradigm''  \citep{futrell-etal-2019-neural}. In our analyses, we compare human reading times to PLM surprisal estimates. Surprisal is the negative log probability of a word in context and has been previously related to human reading times \citep{hale2001probabilistic, levy2008expectation, demberg2008data,smith2013effect} as well as neuropsychological effects such as the N400 \citep{frank2015erp,kutas1989electrophysiological}, which represent human processing difficulty.
We collect a new dataset, Causality in Script Knowledge (\textbf{CSK}), consisting of short stories about daily activities which are typically part of the \textit{script knowledge} of humans, see Figure \ref{fig:example_materials} for an example. The term ``script knowledge'' refers to commonsense knowledge about everyday activities, where ``scripts'' are defined as prototypical sequences of events in these activities. The stories are constructed such that they contain a pair of events, $A$ and $B$ which are causally contingent on one another. We manipulate event $A$ to be stated, negated or omitted, and subsequently measure processing difficulty on event $B$. 

Our first research question (\textbf{RQ1}) relates to the effect of the incoherence in the $\neg A \rightarrow B$ condition, compared to the coherent $A \rightarrow B$ condition. For humans, a large body of previous literature  \citep{bloom1990line, radvansky2014different,singer1996role} leads us to expect that human readers will notice the inconsistency and that this can be measured in terms of slower reading times on event $B$. For language models, we want to test whether and which models also exhibit a similar effect, by comparing the surprisal values for the words of event $B$ following the $A$ vs.~$\neg A$ mentioned in the previous context. In order for a language model to handle this case, it needs to (a) understand the contingency between events A and B (even though they systematically don't use overlapping lexical items) and (b) be able to represent event $A$ or $\neg A$ effectively across the intervening sentences so it is still represented when encountering $B$. We find that the large models (GPT-3 and Vicuna) do well on this task, but smaller models mostly fail.

Our second research question (\textbf{RQ2}) aims to tap into how script knowledge facilitates language comprehension. To this end, we compare the processing of event $B$ in a setting where neither event $A$ nor event $\neg A$ are mentioned in the previous context. If comprehenders integrate their script knowledge with the text, they should have an easy time processing event $B$ even without the prior mention of event $A$ \citep{bower1979scripts}. The previous literature on human sentence processing has no direct evidence about the processing difficulty of event B in this case, so here our experiment makes a new contribution: we find that humans are significantly faster in reading segment $B$ in the $nil \rightarrow B$ condition compared to $\neg A \rightarrow B$, and that reading times between conditions $nil \rightarrow B$ and $A \rightarrow B$ do not differ significantly from one another. 
Our subsequent evaluation of LLMs on the same contrast however shows that all LLMs fail to show human-like processing: they do not have lower surprisal on the $nil \rightarrow B$ condition than on $\neg A \rightarrow B$ -- some models even assign higher surprisal estimates to the $nil \rightarrow B$ condition, indicating that even the most recent large LLMs in our evaluation cannot effectively integrate script knowledge for estimating the probability of upcoming words.


\section{Background} \label{Related Work}
\subsection{Causal inference and script knowledge}


When humans read text, they connect events mentioned in the text into a locally and globally coherent causal network, thereby not only integrating information from the text but also based on context and commonsense knowledge
\citep{van1990causal, graesser1997discourse}. 
It has been shown that when the causal network does not support new events or the new event contradicts the previous text, readers experience processing difficulties, resulting in longer reading times \citep{bloom1990line, radvansky2014different}. 
%
%
The comprehension of a new event also relies on commonsense knowledge \citep{hare2009activating}. In fact, \citet{singer1996role} showed that when commonsense knowledge does not support an event described in the text, comprehenders take more time processing it.

A special type of commonsense knowledge that was shown to also modulate reading comprehension is script knowledge \citep{abbott1985representation, bower1979scripts, schank1975structure}. Scripts represent knowledge structures consistent with sets of beliefs built on past experiences about everyday, routine, and conventional activities like baking a cake.
Importantly, the events constituting a script can be highly causally inter-connected and are crystallized in memory -- one can expect script-related events to be activated once the script is invoked. In a series of experiments, \citet{bower1979scripts} showed that after subjects read an everyday story that constituted a script, they also recalled script-related events that were not explicitly mentioned in the story \citep[see][for similar findings showing that script knowledge is an indistinguishable part of the memory representation]{gibbs1980concept}. In turn, it is expected that when reading a story, script-related events can be primed by the script itself rather than by some single events mentioned in the text, without processing time loss \citep{keenan1974identification}.


\subsection{Experiments with language models}

\myparagraph{Causal Reasoning.} Recent LLMs such as GPT-3.5 \citep{neelakantan2022gpt3.5} have achieved strong performance in many reasoning tasks under zero-shot settings, such as symbolic reasoning, logical reasoning, mathematical reasoning and commonsense inference \citep{kojima2022zeroshotcot}. The common practice to conduct zero-shot reasoning is \textit{prompting}, i.e.~to append a task-specific text to the input to LLMs and then sample the output \citep{radford2019language}. Although the cause is usually provided in the prompt (like condition $A \rightarrow B$), LLMs can reason without relying only on surface cues like word overlap \citep{lampinen-etal-2022-language}. 
On top of that, LLMs can be prompted to produce explicit reasoning steps with chain-of-thought prompting \citep{wei2022cot}. 

\myparagraph{Script knowledge.} 
Recent studies have suggested that LLMs may learn script knowledge as part of their training \citep{sakaguchi-etal-2021-proscript-partially,sancheti-rudinger-2022-large}. 
\citet{ravi-etal-2023-happens} fine-tune GPT-3 to automatically generate plausible events that happen before and after a given event, and  \citet{yuan2023distilling} report promising results on prompting an InstructGPT model \citep{ouyang2022instructgpt} to automatically generate scripts and then filtering results in the second step. Similarly, \citet{brahman2023plasma} use a distilled small LM as script planner and fine-tuned RoBERTa as verifiers. 

There are however also reports that indicate that script knowledge in LLMs may not yet be sufficient: zero-shot probing on GPT-2 has been found to generate poor event sequences \citep{sancheti-rudinger-2022-large},
and GPT-3 was found to be only marginally better than chance on predicting event likelihoods \citep{zhang-etal-2023-causal} and exhibit poor performance on event temporal ordering \citep{suzgun-etal-2023-challenging}. 


Several ways of specifically integrating commonsense knowledge into LLMs have been proposed: some LLMs are trained from scratch on structural data with commonsense knowledge like knowledge graphs~\citep[ERNIE;][]{zhang-etal-2019-ernie} and semantic frames~\citep[SpanBERT;][]{joshi-etal-2020-spanbert}. \citet{bosselut-etal-2019-comet,hwang2021cometatomic} further equips LLMs with structural input and output to model commonsense knowledge. 
In the present contribution, we explore previous models that have been reported to be successful on inference tasks.
More details of the choice of LLMs are in Section \ref{sec:lm-choice}. 

\subsection{The TRIP dataset}
\label{sec:trip} 
A dataset that is particularly relevant to the present study is the TRIP dataset, which contains 1472 pairs of two similar stories, which differ by one sentence at a ``breakpoint'' position \citep{storks-etal-2021-tiered-reasoning}. One of the stories is plausible, and the other one is implausible, due to a causal conflict between the sentence at the breakpoint position and an earlier part of the text. 
The plausible stories correspond to the $A \rightarrow B$ condition in our dataset, while the implausible stories correspond to our $\neg A \rightarrow B$ condition. The  breakpoint sentence corresponds to our critical sequence~$B$. 

\citet{richardson-etal-2022-breakpoint} fine-tune a T5 model augmented with logical states (BPT) of each event to detect the causal conflicts and outperform a RoBERTa baseline by a large margin. \citet{ma-etal-2022-coalescing} fine-tune a framework to integrate global and local information, which further outperforms BPT. Our aim is not to finetune the LLMs on TRIP but test them in a zero-shot fashion.

\section{Experiments with Humans}

\subsection{Dataset}
\label{sec:dataset}
The Causality in Script Knowledge (\textbf{CSK}) data set consists of 21 English stories describing everyday activities, 
built on top of DeScript dataset \cite{wanzare-etal-2016-crowdsourced} -- see Figure \ref{fig:example_materials}, part I. 
Each story starts with a script initiation (e.g., ``she decided to bake a cake'') -- thus, readers can already activate script knowledge about the event at that point. A pair of events A and B represent our main interest. They were chosen in such a way that event A (``get the cake decorations'') enabled the happening of event B (``add star-shaped sprinkles''). Importantly, no other events in the story draw a direct causal link to event B, except event A and the script itself. Events A and B were always separated by other intermediate script events (73.6 words on average; $sd = 10.3$; min: 59; max: 91).
The descriptive statistics for the stories are presented in Appendix~\ref{app:descriptive_materials}.


\subsection{Experimental conditions}
Our target manipulation relates to the appearance of events A and B in the story thus producing three different story conditions:

\myparagraph{$A \rightarrow B$.} \label{AandB} Event B logically follows event A within the story context. In this way, event A draws a direct causal link to event B, and thus event B is anticipated to happen on the basis of event A. 

\myparagraph{$\neg A \rightarrow B$.} \label{notAandB}
Event A is explicitly negated, making the occurrence of event B implausible or even impossible. The mention of event B thus is unexpected and stands in a causal conflict with the earlier information.

\myparagraph{$nil \rightarrow B$.} \label{onlyB}
Event A is omitted. Even though event A is not explicitly stated, it is expected that humans will easily infer its occurrence from the context, making the mention of event B plausible and easy to integrate \cite{bower1979scripts}.

\subsection{Experimental procedure}
For data collection, each story was divided into paragraphs or text chunks (as shown, for example, in Figure \ref{fig:example_materials}, part II). 
During the experiment, subjects saw only one paragraph at a time (chunk-by-chunk presentation). 
After reading each story, subjects had to rate how sure they were about the events A and B to have occurred, on a Likert scale ranging from 0 (\textit{Not sure at all}) to 7 (\textit{Very sure}) --  see Figure \ref{fig:example_materials}, part III.
To measure the processing difficulties of humans, we compare the reading times for event $B$ across the experimental conditions. More details about subjects' belief ratings are presented in Appendix \ref{results:beliefs}.

251 native English speakers were hired via a crowdsourcing platform Prolific\footnote{\url{https://www.prolific.co/}} to participate in the study. Each participant read three stories. Each story had a different topic and was presented in a different condition. 

\subsection{Analysis}
To investigate the effects of processing difficulty that event B causes in subjects depending on a story condition, we analyse mean per character reading times associated with the chunks that contain event B. The log-transformed reading times were analysed using linear mixed-effects regression models \citep[LMER;][]{bates2014fitting}. The maximal random effects structure included by-subject and by-item random intercepts and by-item random slopes for story condition and was simplified for convergence when needed. 

\begin{figure}[t]
    \centering
    \includegraphics[width=0.99\columnwidth]{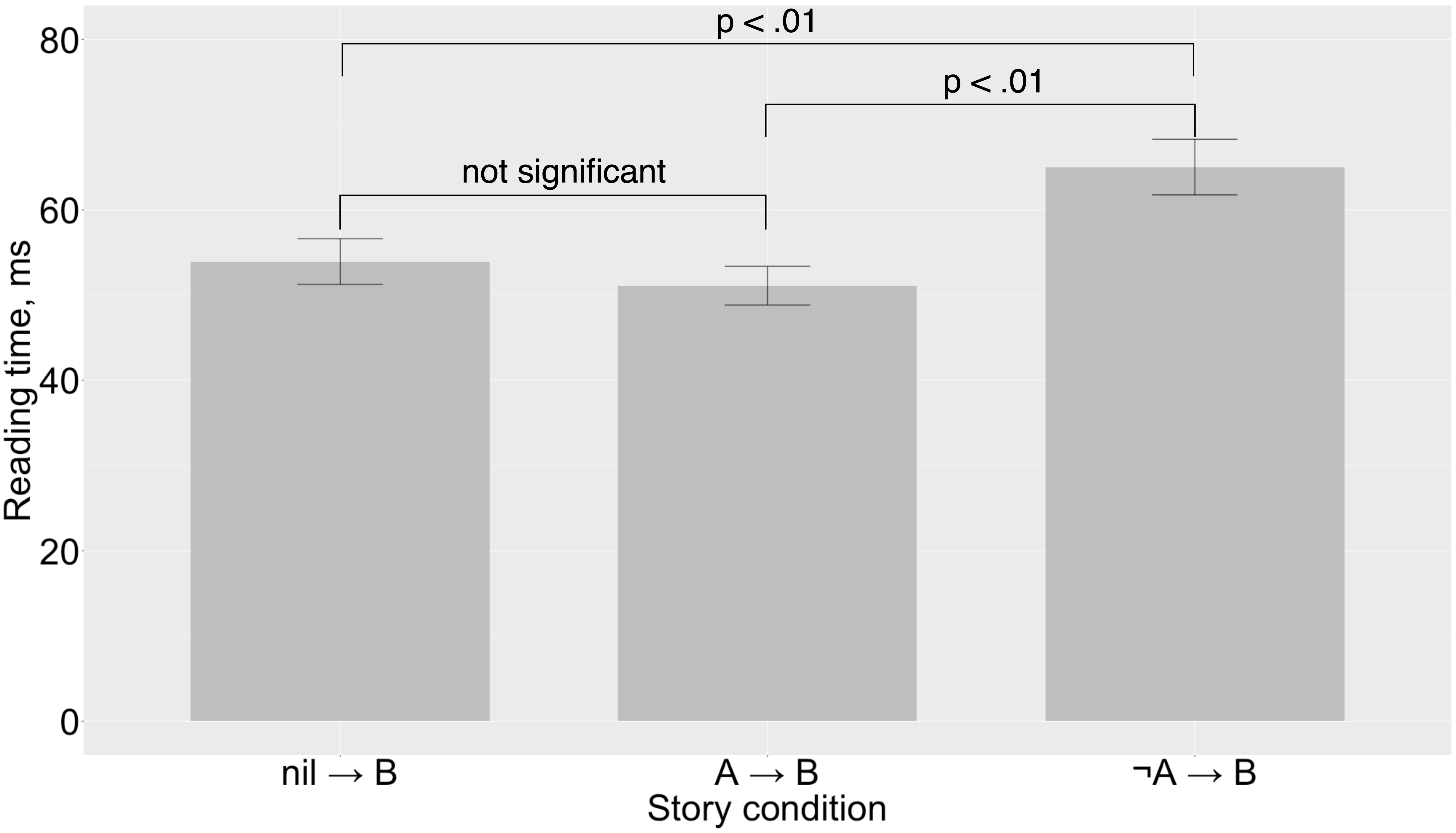}
    \caption{Mean by-character reading times at event $B$,  by story condition; p-values are taken from the corresponding LMER models, see Section \ref{results:rt}.} 
    \label{fig:RTcharB}
\end{figure}

\subsection{Results}
\label{results:rt}

To answer to what extent causal inconsistencies are reflected in human language processing (RQ1), we compared reading times on segment $B$ in the $A \rightarrow B$ vs.~$\neg A\rightarrow B$ conditions. The random effects structure included by-subject and by-item random intercepts and by-item random slopes for story conditions. We found that subjects read chunks with event $B$ significantly more slowly when event $A$ was explicitly negated in the story ($b=0.21$, $se = 0.04$, $t=4.77$, $p<.01$), see also Figure \ref{fig:RTcharB}.



To analyse subjects' ability to infer causal links from script knowledge (RQ2), we compared the reading times in $nil\rightarrow B$ vs.~$A\rightarrow B$ conditions. The random effects structure included by-item random intercepts. We observed no significant difference between these conditions ($b=-0.04$, $se = 0.05$, $t=-0.7$, $p=.48$). Thus, the absence of event A, which serves as a direct causal link to event B, does not slow event's B processing in terms of reading times. Note that the reading time of condition $\neg A \rightarrow B$ is significantly slower than the reading time in condition $nil \rightarrow B$ ($b=0.17$, $se = 0.05$, $t=3.23$, $p < .01$). 



\begin{table*}[t]
\centering
\resizebox{0.65\textwidth}{!}{%
\begin{tabular}{|c|c||c|c|c|c|c|c||c|}
\hline
Story ID & Condition & added & star & - & shaped & sprink & les & avg. S \\
\hline
2 & $A \rightarrow B$ & 0.09 & 0.00       & 0.95 & 1.00 & 0.97 & 1.00 & 20.31 \\
2 & $nil \rightarrow B$ & 0.02 & 0.00       & 0.97 & 1.00 & 0.96 & 1.00 & 27.94 \\
2 & $\neg A \rightarrow B$ & 0.01 & 0.00 & 0.98 & 1.00 & 0.98 & 1.00 & 27.98 \\
\hline
\end{tabular}
}
\caption{Sample raw data of the baking a cake story from GPT-3, showing the rounded probabilities assigned by the language model to each token in a sequence of story number 1 under three different conditions ($A \rightarrow B$, $nil \rightarrow B$, and $\neg A \rightarrow B$). The last column represents the average surprisal (avg. S) for the corresponding sequence.}
\label{table:GPT-3 example}
\end{table*}

\begin{table*}[ht]
\centering
\resizebox{0.99\textwidth}{!}{%
\begin{tabular}{lr|rrr|rrr|rrr}
\textbf{Model Name} & \# para. & $b$ & $t$ & sign & $b$ & $t$ & sign & $b$ & $t$ & sign \\
& \textbf{(M)} & \multicolumn{3}{c|}{CSK} & \multicolumn{3}{c|}{CSK (short dist)} & \multicolumn{3}{c}{TRIP}  \\
\hline
\multicolumn{2}{c|}{CLM} & & & & & &  \\
GPT-3.5: text-davinci-003 & 175K & 0.59 & 5.87 & *** & 0.20 & 1.59 & n.s. & 0.30 & 10.82 & *** \\
GPT-3.5: text-davinci-002 & 175K & 0.51 & 2.75 & * & 0.10 & 0.70 & n.s. & 0.26 & 7.41 & *** \\
InstructGPT: text-davinci-001 & 175K & 0.26 & 2.03 & . & -0.02 & -0.18 & n.s. & 0.29 & 5.81 & *** \\
InstructGPT: davinci-instruct-beta & 175K & 0.21 & 2.76 & * & 0.12 & 1.78 & . & 0.20 & 8.68 & *** \\
GPT-3: davinci & 175K & 0.21 & 2.76 & * & 0.19 & 2.69 & * & 0.20 & 8.25 & *** \\
Vicuna-13B & 13016 & 0.22 & 2.25 & * & -0.01 & -0.07 & n.s. & 0.26 & 7.56 & *** \\
Vicuna-7B & 6738 & 0.28 & 2.56 & * & 0.12 & 1.08 & n.s. & 0.22 & 6.35 & *** \\
InstructGPT: text-curie-001 & 6700 & 0.03 & 0.31 & n.s. & 0.06 & 0.56 & n.s. & 0.19 & 5.78 & *** \\
GPT-3: curie & 6700 & 0.23 & 3.43 & ** & 0.21 & 3.75 & ** & 0.12 & 5.92 & *** \\
GPT-2: XL & 1638 & 0.05 & 0.96 & n.s. & 0.08 & 1.54 & n.s. & 0.06 & 3.15 & ** \\
GPT-2: L & 838 & 0.04 & 0.77 & n.s. & 0.04 & 0.64 & n.s. & 0.05 & 2.77 & ** \\
XGLM & 827 & -0.03 & -0.79 & n.s. & 0.02 & 0.38 & n.s. & 0.02 & 1.38 & n.s. \\
Bigbird-pegasus-large-arxiv & 470 & 0.06 & 1.20 & n.s. & 0.00 & -0.04 & n.s. & 0.00 & -0.02 & n.s. \\
Pegasus-large & 467 & 0.02 & 0.85 & n.s. & 0.00 & 0.00 & n.s. & 0.00 & -0.48 & n.s. \\
XLNet-large-cased & 393 & -0.03 & -1.99 & . & -0.04 & -2.42 & * & 0.00 & 0.66 & n.s. \\
OPT & 357 & 0.01 & 0.12 & n.s. & 0.02 & 0.32 & n.s. & 0.03 & 1.78 & . \\
GPT-Neo & 164 & 0.03 & 0.67 & n.s. & 0.05 & 1.11 & n.s. & 0.01 & 0.90 & n.s. \\
GPT-2 & 163 & 0.00 & -0.10 & n.s. & 0.03 & 0.74 & n.s. & 0.01 & 0.53 & n.s. \\
GPT: openai-gpt & 148 & 0.00 & -0.01 & n.s. & 0.06 & 1.35 & n.s. & 0.05 & 3.18 & ** \\
\hline
\multicolumn{2}{c|}{MLM} & & & & & &  \\
Bigbird-roberta-large & 412 & 0.18 & 1.64 & n.s. & 0.33 & 1.72 & n.s. & 0.04 & 2.90 & ** \\
Perceiver & 201 & -0.02 & -0.51 & n.s. & 0.04 & 0.79 & n.s. & 0.01 & 1.29 & n.s. \\
Bigbird-roberta-base & 167 & 0.05 & 0.34 & n.s. & -0.03 & -0.13 & n.s. & 0.03 & 2.62 & ** \\
Nystromformer-512 & 132 & 0.06 & 1.50 & n.s. & 0.04 & 0.80 & n.s. & -0.01 & -0.46 & n.s. \\
FNet-base & 108 & 0.01 & 0.14 & n.s. & 0.02 & 0.41 & n.s. & -0.01 & -0.80 & n.s. \\
\hline
\end{tabular}
}
\caption{Results for RQ1 ($A \rightarrow B$ versus $\neg A \rightarrow B$) on CSK (original and intervention removal) and TRIP dataset. The \# para. (M) column shows the number of parameters in millions. n.s. represent that the results are not statistically significant. The ., *, **, and *** in the sign column represent $p$-values < 0.1, 0.05, 0.01, and 0.001. }
\label{tab:CSK_RQ1}
\end{table*}

\section{Can LLMs Detect Causal Conflicts (RQ1)?}
In this section, we measure the ability of different LLMs to track event contingency. 
We feed the script stories into the language models and record the LM's surprisal scores on a word-by-word basis. We then test whether the mean surprisal scores for the critical region (event $B$) differ between conditions. As the script stories corpus is relatively small, we additionally test the models on the TRIP dataset \citep{storks-etal-2021-tiered-reasoning} to assess their recognition of causal incongruencies on a wider set of materials (see Section \ref{sec:trip-expt}).


\subsection{Choices of LLMs}
\label{sec:lm-choice}
We select a set of more than 30 LLMs including both causal language models (CLMs) and masked language models (MLMs) because we want to see the effects of these two loss functions. 


\paragraph{CLMs.}  We first select CLMs because they work in a left-to-right fashion, similar to how the human readers in our experiment read the experimental materials. We chose GPT-1/2/3 and Instruct GPT \citep{radford2018gpt1,radford2019language,brown2020gpt3,ouyang2022instructgpt} models to represent CLMs because these models have been showing the highest performance on many NLP tasks \citep{chang2023language}. We choose GPT-3.5 \citep{neelakantan2022gpt3.5} because it was trained with both programming code and text which demonstrates strong performance on entity tracking \citep{kim2023entity}, a prerequisite for causal reasoning. 
Notably, ChatGPT \citep{chatgpt} and GPT-4 \citep{openai2023gpt4} can not be used with our methods, because the API does not allow access to the probabilities. 
Additionally, we use Vicuna models \citep{vicuna2023} as an approximation to ChatGPT, which is a fine-tuned LLaMA model \citep{touvron2023llama} trained on 70K user-shared ChatGPT conversations. Open models like Vicuna have the advantage of results being reproducible. 
Similarly, we choose OPT \citep{zhang2022opt} and GPT-Neo \citep{gpt-neo} as open versions of GPT-3. 

We also selected task-specific models that could potentially capture script knowledge via exposure to more diverse datasets like summarization models, Pegasus \citep{zhang2020pegasus}, Bigbird-pegasus, and a multilingual model XGLM \citep{lin-etal-2022-shot}. Lastly, we chose XLNet because it has been previously shown to be effective for zero-shot script parsing \citep{zhai-etal-2021-script,zhai-etal-2022-zero} wrt.~handling causal inferences in commonsense stories in a zero-shot setting. 

\paragraph{MLMs.} MLMs are another group of language models that obtained state-of-the-art performances across many NLP tasks. We note that the way they work is not similar to human language processing, and the surprisal estimates obtained from them are not directly comparable to surprisals obtained from left-to-right models. 
However, we decided to include some MLMs that have been specifically designed to handle long-distance dependencies (via their efficient self-attention mechanisms) into our evaluation, to observe how these models perform regarding the causal inferences given long commonsense stories. 
Specifically, we test  Bigbird-roberta \citep{michalopoulos-etal-2022-icdbigbird}, FNet \citep{lee-thorp-etal-2022-fnet}, Nystromformer \citep{xiong2021nystromformer} and Perceiver \citep{jaegle2021perceiver}. 


All models used here were available through either HuggingFace models or the OpenAI API. More details are in Appendix \ref{app:models}, where we briefly describe all the models.

\subsection{Method}
\label{sec:lm-method}
For causal language models (CLMs), we perform word-by-word next-word prediction for event $B$, recording the next token probabilities for each token in segment $B$. 
For masked language models (MLMs), we follow \citet{salazar-etal-2020-masked} to provide models with the context before and after the target token in segment $B$. The pertinent token itself is masked, forcing the masked language models to infer it based on the surrounding context. 
For instance, in the example story in Figure \ref{fig:example_materials}, the words ``added star-shaped sprinkles'' constitute the target region describing event $B$. Each token in this sequence was masked one at a time. We then calculated the probabilities of the masked tokens given the surrounding story context. MLM models thus have more information than CLM models due to the additional information from other tokens in the event $B$ and the context after event $B$. We therefore would like to point out that this method is not cognitively plausible, and that the surprisal scores obtained from them hence will also reflect this ``privileged'' knowledge.
We also note that the surprisal estimation from MLMs can in principle be adapted to simulate left-to-right processing better, but think that this is only worthwhile to explore in more detail if MLMs prove to be successful at modelling the long-distance dependencies relevant to our texts. 

Based on the probability of the target words $w$ given the story context, we then calculate the target tokens' surprisal as their negative log probability: $\mathrm{surprisal}(w) = -\log P(w|\mathrm{story\_context})$.
We then calculate the average per-word surprisal by averaging the surprisal of each word into an estimate of the surprisal of the critical region for each item.
Table \ref{table:GPT-3 example} provides an example of the resulting data format, depicting the probabilities for each token in B and the 
average surprisal values for a given story across all conditions in GPT-3.


\subsection{Data Analysis}
\label{sec:lm-analysis}
To identify the PLM(s) that show comparable effects to humans, we run an equivalent analysis to how the reading time data were analysed: we estimate linear mixed effects models with surprisal as a response variable and condition ($A \rightarrow B$ ~vs. $\neg A \rightarrow B$) as a predictor. The model also includes by-item random intercepts. The formula is: $log(\mathrm{surprisal}) \sim \mathrm{story\_condition} + (1|\mathrm{story})$.

\subsection{Results}


Table \ref{tab:CSK_RQ1} (column named CSK) presents the results for all language models on whether model surprisals were significantly higher for the $\neg A \rightarrow B$ condition than in the $A \rightarrow B$ condition, indicating that the model's surprisal scores reflect the incoherence (RQ1). 
High positive $b$ values indicate that surprisal values are higher on segment $B$ in the $\neg A \rightarrow B$ condition compared to the $A \rightarrow B$ condition. Significance stars indicate whether the differences were statistically reliable. Our results show that none of the MLM models, and only some of the largest CLM models showed a reliable difference in surprisal estimates between the coherent and the incoherent ($\neg A \rightarrow B$) condition. 

GPT-3.5: text-davinci-003 shows the largest effect and high statistical reliability. Further models that show the expected behaviour include other versions of GPT-3/GPT-3.5 and the Vicuna model. Surprisingly, InstructGPT models that are trained with human-selected samples don't show significant effects. This result implies additional training on high-quality samples harms the models' ability to identify causal conflicts. 

\begin{table*}[t]
\centering
\tabcolsep=3pt
\begin{tabular}{l|rrr|rrr}
 & \multicolumn{3}{c}{$nil$ vs.~$\neg A$} & \multicolumn{3}{c}{$nil$ vs.~$A$}  \\
\textbf{Model Name (CLMs only)} &$b$ & $t$ & sign & $b$ & $t$ & sign \\
\hline
GPT-3.5: text-davinci-003 &  0.08 & 0.77 & n.s. & -0.52 & -5.10 & *** \\
GPT-3.5: text-davinci-002 &  -0.06 & -0.38 & n.s. & -0.57 & -3.65 & *** \\
InstrGPT: davinci-instr-beta &  -0.17 & -1.96 & . & -0.39 & -4.36 & *** \\
GPT-3: davinci &  -0.15 & -1.79 & . & -0.36 & -4.34 & *** \\
Vicuna-13B &  -0.15 & -1.52 & n.s. & -0.37 & -3.73 & *** \\
Vicuna-7B &  -0.07 & -0.58 & n.s. & -0.36 & -2.91 & ** \\
GPT-3: curie &  -0.23 & -2.74 & ** & -0.46 & -5.54 & *** \\

\hline
Human & 0.17 & 3.23 & ** & -0.04 & -0.7 & n.s.  \\
\hline
\end{tabular}
\caption{Results for RQ2 ($nil \rightarrow B$ versus $\neg A\rightarrow B$ and $A\rightarrow B$) on CSK dataset. Note that coefficient estimates for human data refer to log reading times, and are hence not directly comparable to the numbers in the CLMs, which estimate the surprisal effect. 
}
\label{tab:RQ2}
\end{table*}

\subsection{Effect of dependency length (distance between events A and B)}
Next, we wanted to check whether the failure of the models that don't show a significant difference between conditions is due to problems with encoding the text effectively and ``remembering'' event $A$ or $\neg A$ when processing event $B$, or whether it is related to failure to detect the mismatch between the events.
We therefore modified the original experiment's design by reducing the distance between events A and B in the story by removing all intervening sentences. (Note that we did not ensure that the removed sentences did not contain crucial information that would compromise the coherence of the story.)


If model failure on the previous task is due to difficulty in handling a long intervening context, we expect that models would show a significant difference between surprisal estimates in this short-distance condition. 


As shown in Table \ref{tab:CSK_RQ1} column named ``CSK (short dist)'', we find that most models show the same behaviour in the short-distance condition and the long-distance condition. 
Interestingly, the results of both GPT-3.5 and Vicuna are non-significant in this condition. This could be due to the removal of intermediary materials, thereby potentially interrupting the causal chains and adversely affecting the activation of event $B$. 
Other models that are still not showing a significant difference between surprisal estimates in the different conditions might be failing due to not recognizing the semantic inconsistency between $\neg A$ and $B$.

\subsection{Experiments on TRIP dataset}
\label{sec:trip-expt}
As the CSK dataset, for which we collected reading times, is relatively small, we also compared the surprisals of the same set of models on the substantially larger TRIP dataset (cf.~Section \ref{sec:trip}), which also contains causal inconsistencies. Their dataset has multiple splits. We do not use the "Order" splits, because that split is too different to our dataset. In those splits, the order of the sentences is switched. We do not use the train splits, as we don't need that many data points. 

We take the stories from the "ClozeDev" split
and run our methods on them and we again estimated surprisal values for each language model, in the same way as described in section \ref{sec:lm-method}. The critical segment $B$ for this dataset corresponds to the breakpoint sentence. The analysis was analogous to the analysis for the CSK dataset.

Table \ref{tab:CSK_RQ1} column ``TRIP'' presents the results of our method on the TRIP dataset. Significant positive effects indicate a significant difference between the model surprisals in the implausible condition compared to the plausible one, indicating that the model recognized the inconsistency correctly. Among the CLMs, GPT-3.5 performs notably well, again displaying the 
largest effect size and p-value $< .001$. 

\subsection{Discussion}
Given the analysis of the CSK and TRIP datasets, we conclude that only some of the GPT models were able to consistently assign higher surprisal to event $B$ (or the breakpoint sentence in TRIP) in the case that causally related event $A$ was explicitly negated earlier in the story. None of the MLMs consistently show this behaviour across the two datasets. Among the GPT models, we find that GPT-3.5: text-davinci-003 shows the most consistent performance. It differs from the others in that it was trained using reinforcement learning from human feedback, which has been found to be correlated with better performance on many reasoning tasks~\citep{chang2023language}.\footnote{We would like to note that we did not apply a correction for multiple testing in the analysis. If we were to more conservatively account for multiple testing, then the results of most models except for GPT-3.5: text-davinci-003 would not be judged as statistically reliable.}

\section{Do LLMs incorporate script knowledge (RQ2)?}
In this section, we are interested in whether the models that can capture the causal link between $A$ and $B$ are also able to integrate script knowledge to a similar extent as humans, i.e.~whether they show a relatively low surprisal even if event $A$ was not explicitly mentioned in the story context. 
We continue with those models showing a significant effect of the $\neg A \rightarrow B$ condition compared to $A \rightarrow B$ 
consistently across the CSK and the TRIP dataset, as these are the only models that seem to reliably capture the causal link.

\begin{figure}[t]
    \centering
    \includegraphics[width=0.99\columnwidth]{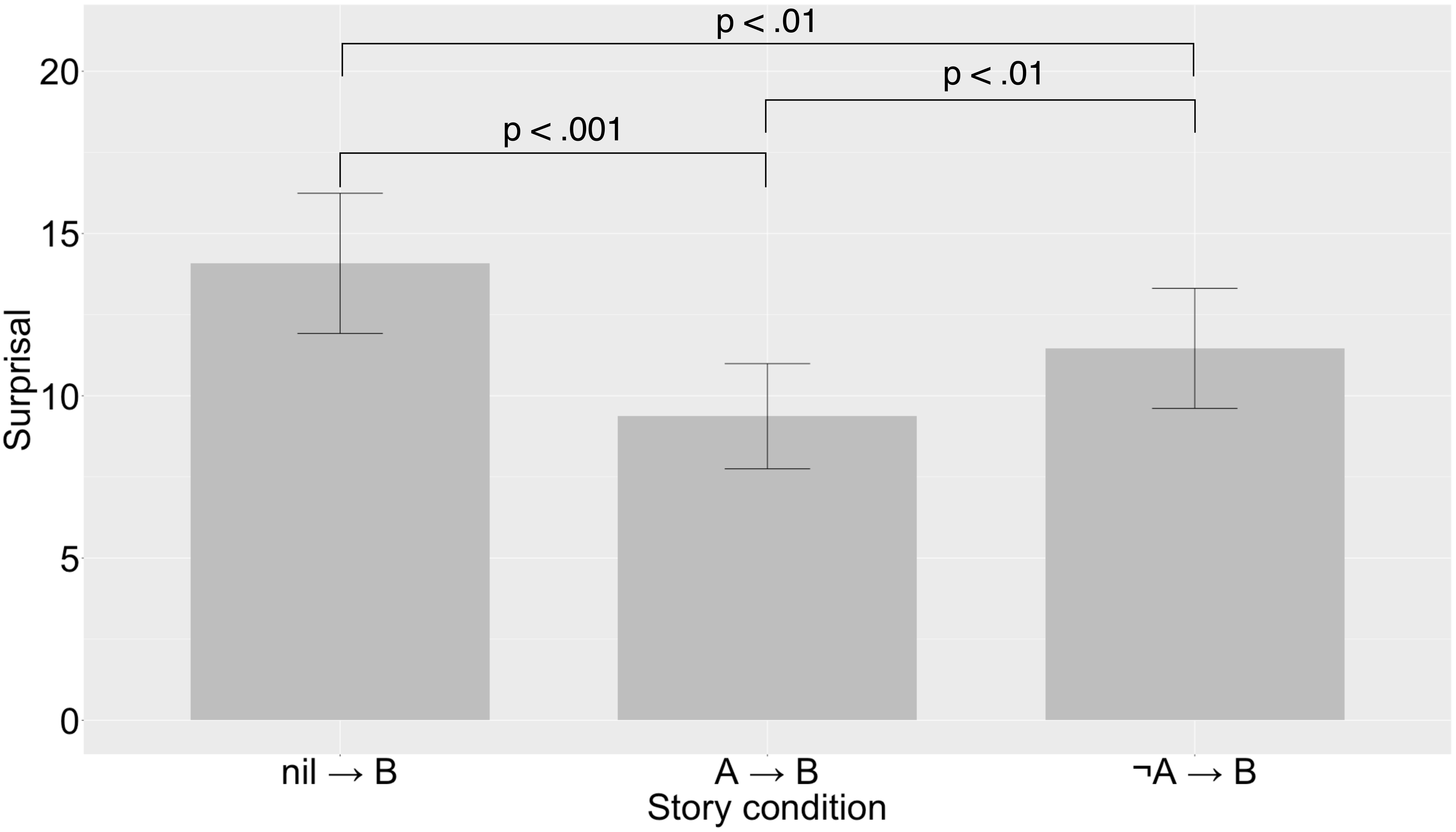}
    \caption{Performance of GPT-3: curie in both research questions. Mean surprisal presented by story condition; p-values are taken from Tables \ref{tab:CSK_RQ1} and \ref{tab:RQ2}} 
    \label{fig:rq2curie}
\end{figure}

\subsection{Analysis and Results}

Analysis was performed using linear mixed-effects models (LMER), similar to Section \ref{sec:lm-analysis}. This time, we compare surprisal estimates of conditions $nil \rightarrow B$ to $\neg A \rightarrow B$ to show firstly whether the model correctly captures the incongruency of $\neg A \rightarrow B$. Next, we compare condition $nil \rightarrow B$ to condition $A \rightarrow B$ in order to determine whether the models are consistent with human readers in terms of NOT showing a large effect. The formula of each LMER model is:
$log(\mathrm{surprisal}) \sim \mathrm{story\_condition} + (1|\mathrm{story})$.

Table \ref{tab:RQ2} shows the results for research question 2. While humans read sequence B is significantly faster in the $nil \rightarrow B$ condition than in the condition with the causal conflict ($\neg A \rightarrow B$), none of the computational models show this effect: most models do not show a significant difference between these conditions, and one model (GPT-3: curie) in fact shows significant effects in the wrong direction ($B$ has higher surprisal in the $nil$ condition than in the $\neg A$ condition), see also Figure \ref{fig:rq2curie}. This might indicate that the lexically related material in condition $\neg A$ (e.g., ``cake decorations'') leads to a relatively low surprisal at region $B$ even if it stands in causal conflict with it. The significantly lower surprisal in condition $A \rightarrow B$ compared to condition $nil \rightarrow B$, which is observed for almost all of the models, furthermore indicates that models fail to include script knowledge effectively in their next word predictions. 

\section{Conclusions}
In this paper, we inspect the behaviors of both large language models and humans in zero-shot causal inference. 
We conducted a self-paced reading experiment on common sense stories to inspect human processing difficulty when reading the stories. Reading time results indicate that humans stumble across causally incoherent text segments, exhibiting longer reading times in these cases. On the other hand, they easily integrate script-predictable information, even if the explicit causal component (event $A$) is missing from the story. 

When we apply the same study to LLMs, only the newest LLMs show similar behavior to humans on encountering casual conflicts. All models fail to replicate human behaviors when the cause is omitted. Even models trained with programming code and instructions fail to make use of script knowledge, which indicates that script knowledge may not be represented sufficiently well in the LLMs tested in this study. We also inspect the effects of the dependency length between the causally linked segments in the text and find that removing the intervention contexts does not improve the models’ performance.


\section{Limitations} 
One limitation from the NLP perspective of our study is that the size of the CSK dataset is small and only in English (only 21 stories). This is a very common limitation of psycholinguistic studies due to the costs of human experiments. We here addressed this shortcoming by also evaluating on the larger dataset TRIP, but a dataset with more stories or more readers would further improve the reliability of the results. Another limitation is that we don't experiment with few-shot examples in prompts, which could have been used to remind the LLMs to make use of script knowledge. We chose the zero-shot setting because humans use script knowledge for casual inference without any ``examples'' and we believe that the LLMs should have the same behaviors as humans. However, this means that our results do not necessarily generalize to other ways of prompting models.
Additionally, we didn't experiment with the most recent OpenAI models like GPT-4 because their official API doesn't support generating the probability output. Lastly, We didn't test models with more than 20B parameters on our own server due to limited hardware resources. We plan to test these models when we have access to more resources. 

Another possible limitation of our experiment is that we cannot comment on the generalizability of our script materials to more general script-based stories for scripts that may be less well-known to human readers. For our materials, we asked participants after each experimental trial whether they were familiar with the script
(``Please tick this box if you have never baked a cake or you have very little experience with it)’’.  Participants answered in 11.2\% of trials that they were not familiar with the script. We observed an effect of familiarity on reading times, showing that subjects read the story faster when they were not familiar with the topic. We note that findings also remained stable when we removed such trials from our analysis. 


\section{Ethics Statement}
The human study was approved by the ethics committee of Deutsche Gesellschaft für Sprachwissenschaft (DGfS). 

\section*{Acknowledgement}
This work was supported by the Deutsche Forschungsgemeinschaft (DFG), Funder Id: http://dx.doi.org/10.13039/501100001659, Project-ID 232722074 -- SFB1102: Information Density and Linguistic Encoding. We sincerely thank the anonymous reviewers for their insightful comments that helped us to improve this paper. We thank Sebastian Schuster and Alexandra Mayn for their informative advice and Nina Shvetsova for her help with the Vicuna models. We also thank students Teresa Martín Soeder and Alice Virginia Chase for helping with materials construction.

\bibliographystyle{acl_natbib}
\bibliography{anthology,custom}

\begin{thebibliography}{66}
\expandafter\ifx\csname natexlab\endcsname\relax\def\natexlab#1{#1}\fi

\bibitem[{Abbott et~al.(1985)Abbott, Black, and
  Smith}]{abbott1985representation}
Valerie Abbott, John~B Black, and Edward~E Smith. 1985.
\newblock The representation of scripts in memory.
\newblock \emph{Journal of memory and language}, 24(2):179--199.

\bibitem[{Bates et~al.(2015)Bates, M{\"a}chler, Bolker, and
  Walker}]{bates2014fitting}
Douglas Bates, Martin M{\"a}chler, Ben Bolker, and Steve Walker. 2015.
\newblock Fitting linear mixed-effects models using lme4.
\newblock \emph{Journal of Statistical Software}, 67:1--48.

\bibitem[{Black et~al.(2021)Black, Gao, Wang, Leahy, and Biderman}]{gpt-neo}
Sid Black, Leo Gao, Phil Wang, Connor Leahy, and Stella Biderman. 2021.
\newblock \href {https://doi.org/10.5281/zenodo.5297715} {{GPT-Neo: Large Scale
  Autoregressive Language Modeling with Mesh-Tensorflow}}.
\newblock \emph{Zenodo}.

\bibitem[{Bloom et~al.(1990)Bloom, Fletcher, Van Den~Broek, Reitz, and
  Shapiro}]{bloom1990line}
Charles~P Bloom, Charles~R Fletcher, Paul Van Den~Broek, Laura Reitz, and
  Brian~P Shapiro. 1990.
\newblock An on-line assessment of causal reasoning during comprehension.
\newblock \emph{Memory \& cognition}, 18:65--71.

\bibitem[{Bosselut et~al.(2019)Bosselut, Rashkin, Sap, Malaviya, Celikyilmaz,
  and Choi}]{bosselut-etal-2019-comet}
Antoine Bosselut, Hannah Rashkin, Maarten Sap, Chaitanya Malaviya, Asli
  Celikyilmaz, and Yejin Choi. 2019.
\newblock \href {https://doi.org/10.18653/v1/P19-1470} {{COMET}: Commonsense
  transformers for automatic knowledge graph construction}.
\newblock In \emph{Proceedings of the 57th Annual Meeting of the Association
  for Computational Linguistics}, pages 4762--4779, Florence, Italy.
  Association for Computational Linguistics.

\bibitem[{Bower et~al.(1979)Bower, Black, and Turner}]{bower1979scripts}
Gordon~H Bower, John~B Black, and Terrence~J Turner. 1979.
\newblock Scripts in memory for text.
\newblock \emph{Cognitive psychology}, 11(2):177--220.

\bibitem[{Brahman et~al.(2023)Brahman, Bhagavatula, Pyatkin, Hwang, Li, Arai,
  Sanyal, Sakaguchi, Ren, and Choi}]{brahman2023plasma}
Faeze Brahman, Chandra Bhagavatula, Valentina Pyatkin, Jena~D Hwang,
  Xiang~Lorraine Li, Hirona~J Arai, Soumya Sanyal, Keisuke Sakaguchi, Xiang
  Ren, and Yejin Choi. 2023.
\newblock Plasma: Making small language models better procedural knowledge
  models for (counterfactual) planning.
\newblock \emph{arXiv preprint arXiv:2305.19472}.

\bibitem[{Brown et~al.(2020)Brown, Mann, Ryder, Subbiah, Kaplan, Dhariwal,
  Neelakantan, Shyam, Sastry, Askell et~al.}]{brown2020gpt3}
Tom Brown, Benjamin Mann, Nick Ryder, Melanie Subbiah, Jared~D Kaplan, Prafulla
  Dhariwal, Arvind Neelakantan, Pranav Shyam, Girish Sastry, Amanda Askell,
  et~al. 2020.
\newblock Language models are few-shot learners.
\newblock \emph{Advances in neural information processing systems},
  33:1877--1901.

\bibitem[{Bubeck et~al.(2023)Bubeck, Chandrasekaran, Eldan, Gehrke, Horvitz,
  Kamar, Lee, Lee, Li, Lundberg et~al.}]{bubeck2023sparks}
S{\'e}bastien Bubeck, Varun Chandrasekaran, Ronen Eldan, Johannes Gehrke, Eric
  Horvitz, Ece Kamar, Peter Lee, Yin~Tat Lee, Yuanzhi Li, Scott Lundberg,
  et~al. 2023.
\newblock Sparks of artificial general intelligence: Early experiments with
  gpt-4.
\newblock \emph{arXiv preprint arXiv:2303.12712}.

\bibitem[{Chang and Bergen(2023)}]{chang2023language}
Tyler~A Chang and Benjamin~K Bergen. 2023.
\newblock Language model behavior: A comprehensive survey.
\newblock \emph{arXiv preprint arXiv:2303.11504}.

\bibitem[{Chiang et~al.(2023)Chiang, Li, Lin, Sheng, Wu, Zhang, Zheng, Zhuang,
  Zhuang, Gonzalez, Stoica, and Xing}]{vicuna2023}
Wei-Lin Chiang, Zhuohan Li, Zi~Lin, Ying Sheng, Zhanghao Wu, Hao Zhang, Lianmin
  Zheng, Siyuan Zhuang, Yonghao Zhuang, Joseph~E. Gonzalez, Ion Stoica, and
  Eric~P. Xing. 2023.
\newblock \href {https://lmsys.org/blog/2023-03-30-vicuna/} {Vicuna: An
  open-source chatbot impressing gpt-4 with 90\%* chatgpt quality}.

\bibitem[{Christensen(2018)}]{christensen2018cumulative}
Rune Haubo~B Christensen. 2018.
\newblock Cumulative link models for ordinal regression with the r package
  ordinal.
\newblock \emph{Submitted in J. Stat. Software}, 35.

\bibitem[{Demberg and Keller(2008)}]{demberg2008data}
Vera Demberg and Frank Keller. 2008.
\newblock Data from eye-tracking corpora as evidence for theories of syntactic
  processing complexity.
\newblock \emph{Cognition}, 109(2):193--210.

\bibitem[{Frank et~al.(2015)Frank, Otten, Galli, and Vigliocco}]{frank2015erp}
Stefan~L Frank, Leun~J Otten, Giulia Galli, and Gabriella Vigliocco. 2015.
\newblock The erp response to the amount of information conveyed by words in
  sentences.
\newblock \emph{Brain and language}, 140:1--11.

\bibitem[{Futrell et~al.(2019)Futrell, Wilcox, Morita, Qian, Ballesteros, and
  Levy}]{futrell-etal-2019-neural}
Richard Futrell, Ethan Wilcox, Takashi Morita, Peng Qian, Miguel Ballesteros,
  and Roger Levy. 2019.
\newblock \href {https://doi.org/10.18653/v1/N19-1004} {Neural language models
  as psycholinguistic subjects: Representations of syntactic state}.
\newblock In \emph{Proceedings of the 2019 Conference of the North {A}merican
  Chapter of the Association for Computational Linguistics: Human Language
  Technologies, Volume 1 (Long and Short Papers)}, pages 32--42, Minneapolis,
  Minnesota. Association for Computational Linguistics.

\bibitem[{Gibbs and Tenney(1980)}]{gibbs1980concept}
Raymond~W Gibbs and Yvette~J Tenney. 1980.
\newblock The concept of scripts in understanding stories.
\newblock \emph{Journal of Psycholinguistic Research}, 9:275--284.

\bibitem[{Graesser et~al.(1997)Graesser, Millis, and
  Zwaan}]{graesser1997discourse}
Arthur~C Graesser, Keith~K Millis, and Rolf~A Zwaan. 1997.
\newblock Discourse comprehension.
\newblock \emph{Annual review of psychology}, 48(1):163--189.

\bibitem[{Graesser et~al.(1994)Graesser, Singer, and
  Trabasso}]{graesser1994constructing}
Arthur~C Graesser, Murray Singer, and Tom Trabasso. 1994.
\newblock Constructing inferences during narrative text comprehension.
\newblock \emph{Psychological review}, 101(3):371.

\bibitem[{Hale(2001)}]{hale2001probabilistic}
John Hale. 2001.
\newblock A probabilistic earley parser as a psycholinguistic model.
\newblock In \emph{Proceedings of the second meeting of the North American
  Chapter of the Association for Computational Linguistics on Language
  technologies}, pages 1--8. Association for Computational Linguistics.

\bibitem[{Hare et~al.(2009)Hare, Jones, Thomson, Kelly, and
  McRae}]{hare2009activating}
Mary Hare, Michael Jones, Caroline Thomson, Sarah Kelly, and Ken McRae. 2009.
\newblock Activating event knowledge.
\newblock \emph{Cognition}, 111(2):151--167.

\bibitem[{Hwang et~al.(2021)Hwang, Bhagavatula, Le~Bras, Da, Sakaguchi,
  Bosselut, and Choi}]{hwang2021cometatomic}
Jena~D Hwang, Chandra Bhagavatula, Ronan Le~Bras, Jeff Da, Keisuke Sakaguchi,
  Antoine Bosselut, and Yejin Choi. 2021.
\newblock (comet-) atomic 2020: On symbolic and neural commonsense knowledge
  graphs.
\newblock In \emph{Proceedings of the AAAI Conference on Artificial
  Intelligence}, volume~35, pages 6384--6392.

\bibitem[{Jaegle et~al.(2022)Jaegle, Borgeaud, Alayrac, Doersch, Ionescu, Ding,
  Koppula, Zoran, Brock, Shelhamer, Henaff, Botvinick, Zisserman, Vinyals, and
  Carreira}]{jaegle2021perceiver}
Andrew Jaegle, Sebastian Borgeaud, Jean-Baptiste Alayrac, Carl Doersch, Catalin
  Ionescu, David Ding, Skanda Koppula, Daniel Zoran, Andrew Brock, Evan
  Shelhamer, Olivier~J Henaff, Matthew Botvinick, Andrew Zisserman, Oriol
  Vinyals, and Joao Carreira. 2022.
\newblock \href {https://openreview.net/forum?id=fILj7WpI-g} {Perceiver {IO}: A
  general architecture for structured inputs \& outputs}.
\newblock In \emph{International Conference on Learning Representations}.

\bibitem[{Joshi et~al.(2020)Joshi, Chen, Liu, Weld, Zettlemoyer, and
  Levy}]{joshi-etal-2020-spanbert}
Mandar Joshi, Danqi Chen, Yinhan Liu, Daniel~S. Weld, Luke Zettlemoyer, and
  Omer Levy. 2020.
\newblock \href {https://doi.org/10.1162/tacl_a_00300} {{S}pan{BERT}: Improving
  pre-training by representing and predicting spans}.
\newblock \emph{Transactions of the Association for Computational Linguistics},
  8:64--77.

\bibitem[{Keenan and Kintsch(1974)}]{keenan1974identification}
JM~Keenan and W~Kintsch. 1974.
\newblock The identification of explicitly and implicitly presented
  information.
\newblock \emph{The representation of meaning in memory}, pages 153--176.

\bibitem[{Kim and Schuster(2023)}]{kim2023entity}
Najoung Kim and Sebastian Schuster. 2023.
\newblock \href {https://doi.org/10.18653/v1/2023.acl-long.213} {Entity
  tracking in language models}.
\newblock In \emph{Proceedings of the 61st Annual Meeting of the Association
  for Computational Linguistics (Volume 1: Long Papers)}, pages 3835--3855,
  Toronto, Canada. Association for Computational Linguistics.

\bibitem[{Kojima et~al.(2022)Kojima, Gu, Reid, Matsuo, and
  Iwasawa}]{kojima2022zeroshotcot}
Takeshi Kojima, Shixiang~Shane Gu, Machel Reid, Yutaka Matsuo, and Yusuke
  Iwasawa. 2022.
\newblock \href {https://openreview.net/forum?id=e2TBb5y0yFf} {Large language
  models are zero-shot reasoners}.
\newblock In \emph{Advances in Neural Information Processing Systems}.

\bibitem[{Kutas and Hillyard(1989)}]{kutas1989electrophysiological}
Marta Kutas and Steven~A Hillyard. 1989.
\newblock An electrophysiological probe of incidental semantic association.
\newblock \emph{Journal of cognitive neuroscience}, 1(1):38--49.

\bibitem[{Lampinen et~al.(2022)Lampinen, Dasgupta, Chan, Mathewson, Tessler,
  Creswell, McClelland, Wang, and Hill}]{lampinen-etal-2022-language}
Andrew Lampinen, Ishita Dasgupta, Stephanie Chan, Kory Mathewson, Mh~Tessler,
  Antonia Creswell, James McClelland, Jane Wang, and Felix Hill. 2022.
\newblock \href {https://aclanthology.org/2022.findings-emnlp.38} {Can language
  models learn from explanations in context?}
\newblock In \emph{Findings of the Association for Computational Linguistics:
  EMNLP 2022}, pages 537--563, Abu Dhabi, United Arab Emirates. Association for
  Computational Linguistics.

\bibitem[{Lee-Thorp et~al.(2022)Lee-Thorp, Ainslie, Eckstein, and
  Ontanon}]{lee-thorp-etal-2022-fnet}
James Lee-Thorp, Joshua Ainslie, Ilya Eckstein, and Santiago Ontanon. 2022.
\newblock \href {https://doi.org/10.18653/v1/2022.naacl-main.319} {{FN}et:
  Mixing tokens with {F}ourier transforms}.
\newblock In \emph{Proceedings of the 2022 Conference of the North American
  Chapter of the Association for Computational Linguistics: Human Language
  Technologies}, pages 4296--4313, Seattle, United States. Association for
  Computational Linguistics.

\bibitem[{Levy(2008)}]{levy2008expectation}
Roger Levy. 2008.
\newblock Expectation-based syntactic comprehension.
\newblock \emph{Cognition}, 106(3):1126--1177.

\bibitem[{Lin et~al.(2022)Lin, Mihaylov, Artetxe, Wang, Chen, Simig, Ott,
  Goyal, Bhosale, Du, Pasunuru, Shleifer, Koura, Chaudhary, O{'}Horo, Wang,
  Zettlemoyer, Kozareva, Diab, Stoyanov, and Li}]{lin-etal-2022-shot}
Xi~Victoria Lin, Todor Mihaylov, Mikel Artetxe, Tianlu Wang, Shuohui Chen,
  Daniel Simig, Myle Ott, Naman Goyal, Shruti Bhosale, Jingfei Du, Ramakanth
  Pasunuru, Sam Shleifer, Punit~Singh Koura, Vishrav Chaudhary, Brian O{'}Horo,
  Jeff Wang, Luke Zettlemoyer, Zornitsa Kozareva, Mona Diab, Veselin Stoyanov,
  and Xian Li. 2022.
\newblock \href {https://aclanthology.org/2022.emnlp-main.616} {Few-shot
  learning with multilingual generative language models}.
\newblock In \emph{Proceedings of the 2022 Conference on Empirical Methods in
  Natural Language Processing}, pages 9019--9052, Abu Dhabi, United Arab
  Emirates. Association for Computational Linguistics.

\bibitem[{Ma et~al.(2022)Ma, Ilievski, Francis, Nyberg, and
  Oltramari}]{ma-etal-2022-coalescing}
Kaixin Ma, Filip Ilievski, Jonathan Francis, Eric Nyberg, and Alessandro
  Oltramari. 2022.
\newblock \href {https://aclanthology.org/2022.coling-1.132} {Coalescing global
  and local information for procedural text understanding}.
\newblock In \emph{Proceedings of the 29th International Conference on
  Computational Linguistics}, pages 1534--1545, Gyeongju, Republic of Korea.
  International Committee on Computational Linguistics.

\bibitem[{Michalopoulos et~al.(2022)Michalopoulos, Malyska, Sahar, Wong, and
  Chen}]{michalopoulos-etal-2022-icdbigbird}
George Michalopoulos, Michal Malyska, Nicola Sahar, Alexander Wong, and Helen
  Chen. 2022.
\newblock \href {https://doi.org/10.18653/v1/2022.bionlp-1.32} {{ICDB}ig{B}ird:
  A contextual embedding model for {ICD} code classification}.
\newblock In \emph{Proceedings of the 21st Workshop on Biomedical Language
  Processing}, pages 330--336, Dublin, Ireland. Association for Computational
  Linguistics.

\bibitem[{Neelakantan et~al.(2022)Neelakantan, Xu, Puri, Radford, Han, Tworek,
  Yuan, Tezak, Kim, Hallacy, Heidecke, Shyam, Power, Nekoul, Sastry, Krueger,
  Schnurr, Such, Hsu, Thompson, Khan, Sherbakov, Jang, Welinder, and
  Weng}]{neelakantan2022gpt3.5}
Arvind Neelakantan, Tao Xu, Raul Puri, Alec Radford, Jesse~Michael Han, Jerry
  Tworek, Qiming Yuan, Nikolas~A. Tezak, Jong~Wook Kim, Chris Hallacy, Johannes
  Heidecke, Pranav Shyam, Boris Power, Tyna~Eloundou Nekoul, Girish Sastry,
  Gretchen Krueger, David~P. Schnurr, Felipe~Petroski Such, Kenny Sai-Kin Hsu,
  Madeleine Thompson, Tabarak Khan, Toki Sherbakov, Joanne Jang, Peter
  Welinder, and Lilian Weng. 2022.
\newblock \href {https://api.semanticscholar.org/CorpusID:246275593} {Text and
  code embeddings by contrastive pre-training}.
\newblock \emph{ArXiv}, abs/2201.10005.

\bibitem[{OpenAI(2022)}]{chatgpt}
OpenAI. 2022.
\newblock \href {https://openai.com/blog/chatgpt} {Introducing chatgpt}.
\newblock \emph{OpenAI Blog}.

\bibitem[{OpenAI(2023)}]{openai2023gpt4}
OpenAI. 2023.
\newblock \href {https://api.semanticscholar.org/CorpusID:257532815} {Gpt-4
  technical report}.
\newblock \emph{ArXiv}, abs/2303.08774.

\bibitem[{Ouyang et~al.(2022)Ouyang, Wu, Jiang, Almeida, Wainwright, Mishkin,
  Zhang, Agarwal, Slama, Ray et~al.}]{ouyang2022instructgpt}
Long Ouyang, Jeffrey Wu, Xu~Jiang, Diogo Almeida, Carroll Wainwright, Pamela
  Mishkin, Chong Zhang, Sandhini Agarwal, Katarina Slama, Alex Ray, et~al.
  2022.
\newblock Training language models to follow instructions with human feedback.
\newblock \emph{Advances in Neural Information Processing Systems},
  35:27730--27744.

\bibitem[{Pearl(2009)}]{pearl2009causality}
Judea Pearl. 2009.
\newblock \emph{Causality}.
\newblock Cambridge university press.

\bibitem[{Radford et~al.(2018)Radford, Narasimhan, Salimans, Sutskever
  et~al.}]{radford2018gpt1}
Alec Radford, Karthik Narasimhan, Tim Salimans, Ilya Sutskever, et~al. 2018.
\newblock Improving language understanding by generative pre-training.
\newblock \emph{OpenAI Blog}.

\bibitem[{Radford et~al.(2019)Radford, Wu, Child, Luan, Amodei, Sutskever
  et~al.}]{radford2019language}
Alec Radford, Jeffrey Wu, Rewon Child, David Luan, Dario Amodei, Ilya
  Sutskever, et~al. 2019.
\newblock Language models are unsupervised multitask learners.
\newblock \emph{OpenAI blog}, 1(8):9.

\bibitem[{Radvansky et~al.(2014)Radvansky, Tamplin, Armendarez, and
  Thompson}]{radvansky2014different}
Gabriel~A Radvansky, Andrea~K Tamplin, Joseph Armendarez, and Alexis~N
  Thompson. 2014.
\newblock Different kinds of causality in event cognition.
\newblock \emph{Discourse Processes}, 51(7):601--618.

\bibitem[{Ravi et~al.(2023)Ravi, Tanner, Ng, and
  Shwartz}]{ravi-etal-2023-happens}
Sahithya Ravi, Chris Tanner, Raymond Ng, and Vered Shwartz. 2023.
\newblock \href {https://aclanthology.org/2023.eacl-main.125} {What happens
  before and after: Multi-event commonsense in event coreference resolution}.
\newblock In \emph{Proceedings of the 17th Conference of the European Chapter
  of the Association for Computational Linguistics}, pages 1708--1724,
  Dubrovnik, Croatia. Association for Computational Linguistics.

\bibitem[{Richardson et~al.(2022)Richardson, Tamari, Sultan, Shahaf, Tsarfaty,
  and Sabharwal}]{richardson-etal-2022-breakpoint}
Kyle Richardson, Ronen Tamari, Oren Sultan, Dafna Shahaf, Reut Tsarfaty, and
  Ashish Sabharwal. 2022.
\newblock \href {https://aclanthology.org/2022.emnlp-main.658} {Breakpoint
  transformers for modeling and tracking intermediate beliefs}.
\newblock In \emph{Proceedings of the 2022 Conference on Empirical Methods in
  Natural Language Processing}, pages 9703--9719, Abu Dhabi, United Arab
  Emirates. Association for Computational Linguistics.

\bibitem[{Sakaguchi et~al.(2021)Sakaguchi, Bhagavatula, Le~Bras, Tandon, Clark,
  and Choi}]{sakaguchi-etal-2021-proscript-partially}
Keisuke Sakaguchi, Chandra Bhagavatula, Ronan Le~Bras, Niket Tandon, Peter
  Clark, and Yejin Choi. 2021.
\newblock \href {https://doi.org/10.18653/v1/2021.findings-emnlp.184}
  {pro{S}cript: Partially ordered scripts generation}.
\newblock In \emph{Findings of the Association for Computational Linguistics:
  EMNLP 2021}, pages 2138--2149, Punta Cana, Dominican Republic. Association
  for Computational Linguistics.

\bibitem[{Salazar et~al.(2020)Salazar, Liang, Nguyen, and
  Kirchhoff}]{salazar-etal-2020-masked}
Julian Salazar, Davis Liang, Toan~Q. Nguyen, and Katrin Kirchhoff. 2020.
\newblock \href {https://doi.org/10.18653/v1/2020.acl-main.240} {Masked
  language model scoring}.
\newblock In \emph{Proceedings of the 58th Annual Meeting of the Association
  for Computational Linguistics}, pages 2699--2712, Online. Association for
  Computational Linguistics.

\bibitem[{Sancheti and Rudinger(2022)}]{sancheti-rudinger-2022-large}
Abhilasha Sancheti and Rachel Rudinger. 2022.
\newblock \href {https://doi.org/10.18653/v1/2022.starsem-1.1} {What do large
  language models learn about scripts?}
\newblock In \emph{Proceedings of the 11th Joint Conference on Lexical and
  Computational Semantics}, pages 1--11, Seattle, Washington. Association for
  Computational Linguistics.

\bibitem[{Schank(1975)}]{schank1975structure}
Roger~C Schank. 1975.
\newblock The structure of episodes in memory.
\newblock In \emph{Representation and understanding}, pages 237--272. Elsevier.

\bibitem[{Singer and Halldorson(1996)}]{singer1996constructing}
Murray Singer and Michael Halldorson. 1996.
\newblock Constructing and validating motive bridging inferences.
\newblock \emph{Cognitive Psychology}, 30(1):1--38.

\bibitem[{Singer and Ritchot(1996)}]{singer1996role}
Murray Singer and Kathryn~FM Ritchot. 1996.
\newblock The role of working memory capacity and knowledge access in text
  inference processing.
\newblock \emph{Memory \& cognition}, 24(6):733--743.

\bibitem[{Smith and Levy(2013)}]{smith2013effect}
Nathaniel~J Smith and Roger Levy. 2013.
\newblock The effect of word predictability on reading time is logarithmic.
\newblock \emph{Cognition}, 128(3):302--319.

\bibitem[{Storks et~al.(2021)Storks, Gao, Zhang, and
  Chai}]{storks-etal-2021-tiered-reasoning}
Shane Storks, Qiaozi Gao, Yichi Zhang, and Joyce Chai. 2021.
\newblock \href {https://doi.org/10.18653/v1/2021.findings-emnlp.422} {Tiered
  reasoning for intuitive physics: Toward verifiable commonsense language
  understanding}.
\newblock In \emph{Findings of the Association for Computational Linguistics:
  EMNLP 2021}, pages 4902--4918, Punta Cana, Dominican Republic. Association
  for Computational Linguistics.

\bibitem[{Suzgun et~al.(2023)Suzgun, Scales, Sch{\"a}rli, Gehrmann, Tay, Chung,
  Chowdhery, Le, Chi, Zhou, and Wei}]{suzgun-etal-2023-challenging}
Mirac Suzgun, Nathan Scales, Nathanael Sch{\"a}rli, Sebastian Gehrmann, Yi~Tay,
  Hyung~Won Chung, Aakanksha Chowdhery, Quoc Le, Ed~Chi, Denny Zhou, and Jason
  Wei. 2023.
\newblock \href {https://doi.org/10.18653/v1/2023.findings-acl.824}
  {Challenging {BIG}-bench tasks and whether chain-of-thought can solve them}.
\newblock In \emph{Findings of the Association for Computational Linguistics:
  ACL 2023}, pages 13003--13051, Toronto, Canada. Association for Computational
  Linguistics.

\bibitem[{Touvron et~al.(2023)Touvron, Lavril, Izacard, Martinet, Lachaux,
  Lacroix, Rozi{\`e}re, Goyal, Hambro, Azhar, Rodriguez, Joulin, Grave, and
  Lample}]{touvron2023llama}
Hugo Touvron, Thibaut Lavril, Gautier Izacard, Xavier Martinet, Marie-Anne
  Lachaux, Timoth{\'e}e Lacroix, Baptiste Rozi{\`e}re, Naman Goyal, Eric
  Hambro, Faisal Azhar, Aurelien Rodriguez, Armand Joulin, Edouard Grave, and
  Guillaume Lample. 2023.
\newblock \href {https://api.semanticscholar.org/CorpusID:257219404} {Llama:
  Open and efficient foundation language models}.
\newblock \emph{ArXiv}, abs/2302.13971.

\bibitem[{Trabasso and Sperry(1985)}]{trabasso1985causal}
Tom Trabasso and Linda~L Sperry. 1985.
\newblock Causal relatedness and importance of story events.
\newblock \emph{Journal of Memory and language}, 24(5):595--611.

\bibitem[{Van~den Broek(1990)}]{van1990causal}
Paul Van~den Broek. 1990.
\newblock The causal inference maker: Towards a process model of inference
  generation in text comprehension.
\newblock \emph{Comprehension processes in reading}, pages 423--445.

\bibitem[{Wang et~al.(2019)Wang, Pruksachatkun, Nangia, Singh, Michael, Hill,
  Levy, and Bowman}]{wang2019superglue}
Alex Wang, Yada Pruksachatkun, Nikita Nangia, Amanpreet Singh, Julian Michael,
  Felix Hill, Omer Levy, and Samuel Bowman. 2019.
\newblock Superglue: A stickier benchmark for general-purpose language
  understanding systems.
\newblock \emph{Advances in neural information processing systems}, 32.

\bibitem[{Wanzare et~al.(2016)Wanzare, Zarcone, Thater, and
  Pinkal}]{wanzare-etal-2016-crowdsourced}
Lilian D.~A. Wanzare, Alessandra Zarcone, Stefan Thater, and Manfred Pinkal.
  2016.
\newblock \href {https://aclanthology.org/L16-1556} {A crowdsourced database of
  event sequence descriptions for the acquisition of high-quality script
  knowledge}.
\newblock In \emph{Proceedings of the Tenth International Conference on
  Language Resources and Evaluation ({LREC}'16)}, pages 3494--3501,
  Portoro{\v{z}}, Slovenia. European Language Resources Association (ELRA).

\bibitem[{Wei et~al.(2022)Wei, Wang, Schuurmans, Bosma, Ichter, Xia, Chi, Le,
  and Zhou}]{wei2022cot}
Jason Wei, Xuezhi Wang, Dale Schuurmans, Maarten Bosma, Brian Ichter, Fei Xia,
  Ed~Chi, Quoc~V Le, and Denny Zhou. 2022.
\newblock \href
  {https://proceedings.neurips.cc/paper_files/paper/2022/file/9d5609613524ecf4f15af0f7b31abca4-Paper-Conference.pdf}
  {Chain-of-thought prompting elicits reasoning in large language models}.
\newblock In \emph{Advances in Neural Information Processing Systems},
  volume~35, pages 24824--24837. Curran Associates, Inc.

\bibitem[{Xiong et~al.(2021)Xiong, Zeng, Chakraborty, Tan, Fung, Li, and
  Singh}]{xiong2021nystromformer}
Yunyang Xiong, Zhanpeng Zeng, Rudrasis Chakraborty, Mingxing Tan, Glenn Fung,
  Yin Li, and Vikas Singh. 2021.
\newblock Nystr{\"o}mformer: A nystr{\"o}m-based algorithm for approximating
  self-attention.
\newblock In \emph{Proceedings of the AAAI Conference on Artificial
  Intelligence}, volume~35, pages 14138--14148.

\bibitem[{Yuan et~al.(2023)Yuan, Chen, Fu, Ge, Shah, Jankowski, Yang, and
  Xiao}]{yuan2023distilling}
Siyu Yuan, Jiangjie Chen, Ziquan Fu, Xuyang Ge, Soham Shah, Charles~Robert
  Jankowski, Deqing Yang, and Yanghua Xiao. 2023.
\newblock Distilling script knowledge from large language models for
  constrained language planning.
\newblock \emph{arXiv preprint arXiv:2305.05252}.

\bibitem[{Zhai et~al.(2022)Zhai, Demberg, and Koller}]{zhai-etal-2022-zero}
Fangzhou Zhai, Vera Demberg, and Alexander Koller. 2022.
\newblock \href {https://aclanthology.org/2022.coling-1.356} {Zero-shot script
  parsing}.
\newblock In \emph{Proceedings of the 29th International Conference on
  Computational Linguistics}, pages 4049--4060, Gyeongju, Republic of Korea.
  International Committee on Computational Linguistics.

\bibitem[{Zhai et~al.(2021)Zhai, {\v{S}}krjanec, and
  Koller}]{zhai-etal-2021-script}
Fangzhou Zhai, Iza {\v{S}}krjanec, and Alexander Koller. 2021.
\newblock \href {https://doi.org/10.18653/v1/2021.starsem-1.18} {Script parsing
  with hierarchical sequence modelling}.
\newblock In \emph{Proceedings of *SEM 2021: The Tenth Joint Conference on
  Lexical and Computational Semantics}, pages 195--201, Online. Association for
  Computational Linguistics.

\bibitem[{Zhang et~al.(2020)Zhang, Zhao, Saleh, and Liu}]{zhang2020pegasus}
Jingqing Zhang, Yao Zhao, Mohammad Saleh, and Peter Liu. 2020.
\newblock Pegasus: Pre-training with extracted gap-sentences for abstractive
  summarization.
\newblock In \emph{International Conference on Machine Learning}, pages
  11328--11339. PMLR.

\bibitem[{Zhang et~al.(2023)Zhang, Xu, Yang, Zhou, You, Arora, and
  Callison-Burch}]{zhang-etal-2023-causal}
Li~Zhang, Hainiu Xu, Yue Yang, Shuyan Zhou, Weiqiu You, Manni Arora, and Chris
  Callison-Burch. 2023.
\newblock \href {https://aclanthology.org/2023.findings-eacl.31} {Causal
  reasoning of entities and events in procedural texts}.
\newblock In \emph{Findings of the Association for Computational Linguistics:
  EACL 2023}, pages 415--431, Dubrovnik, Croatia. Association for Computational
  Linguistics.

\bibitem[{Zhang et~al.(2022)Zhang, Roller, Goyal, Artetxe, Chen, Chen, Dewan,
  Diab, Li, Lin et~al.}]{zhang2022opt}
Susan Zhang, Stephen Roller, Naman Goyal, Mikel Artetxe, Moya Chen, Shuohui
  Chen, Christopher Dewan, Mona Diab, Xian Li, Xi~Victoria Lin, et~al. 2022.
\newblock Opt: Open pre-trained transformer language models.
\newblock \emph{arXiv preprint arXiv:2205.01068}.

\bibitem[{Zhang et~al.(2019)Zhang, Han, Liu, Jiang, Sun, and
  Liu}]{zhang-etal-2019-ernie}
Zhengyan Zhang, Xu~Han, Zhiyuan Liu, Xin Jiang, Maosong Sun, and Qun Liu. 2019.
\newblock \href {https://doi.org/10.18653/v1/P19-1139} {{ERNIE}: Enhanced
  language representation with informative entities}.
\newblock In \emph{Proceedings of the 57th Annual Meeting of the Association
  for Computational Linguistics}, pages 1441--1451, Florence, Italy.
  Association for Computational Linguistics.

\end{thebibliography}

\appendix
\label{sec:apppendix}
\section{Experimental materials}
\label{app:descriptive_materials}

When constructing the experimental materials, we controlled for the following parameters: the number of words and text chunks in a story, the number of text chunks and words between events A and B, the number of words in the text chunks that contained event B, and number of words in the chunk after the chunk with event B. The full list of descriptive statistics for our materials is presented in Table \ref{tab:descriptive}.

Each story starts with script initiation -- a sentence in the first chunk that introduces the topic to the reader, e.g., ``\textit{Yesterday Anne had a party at her house, so \textbf{she decided to bake a cake}.}'' from Figure \ref{fig:example_materials}. Sequences of script-related events were built on top of \citet{wanzare-etal-2016-crowdsourced}. Script-related events A and B were chosen in such a way that event A (get the cake decorations) enabled the occurrence of event B (add star-shaped sprinkles). There were no other events in the story that are causally linked to event B. Finally, the chunk with event B always consists of one sentence with the following structure: ``\textit{ADVERB PERSON X did action B and then did a subsequent action from the script sequence.}'' (except the laundry story, where the sentence started with ``She'').

Prior to the analysis, we removed all trials related to the bowling story item, due to a typo. Further, we removed trials where the reading times in the chunk containing event $B$ were shorter than 100ms or larger than 50s. 704 trials from 251 subjects (73\% female; mean age = 40, sd = 14.6, [18;80] range) were available for analysis (1.81\% data loss).

\begin{table}[t]
    \begin{center} 
        \begin{tabular}{lrcrrr} 
        \hline\hline
        parameter & mean & sd \\
        \hline\hline
        $\#$ of words in story:         & & \\
        $A \rightarrow B$      & 158.2  & 12  \\
        $\neg A \rightarrow B$ & 159.1  & 14  \\
        $nil \rightarrow B$    &  150.1 & 11.7 \\
        \hline
        $\#$ of text chunks in story & 6.8 & 0.77 \\
        \hline
        $\#$ of words in chunk with A & 27.6 & 11.3 \\
        \hline
        $\#$ of words in chunk with $\neg A$ & 29.3 & 13.1 \\
        \hline
        $\#$ of words in chunk with B & 12.9 & 1.7 \\
        \hline
        $\#$ of words in chunk after B & 12.9 & 1.8 \\
        \hline
        $\#$ of words b/w A and B: & &  \\
        $A \rightarrow B$      & 73.6 & 10.3  \\
        $\neg A \rightarrow B$ & 71.8 & 12.9  \\
        \hline
        $\#$ of words in A & 7.3 & 3.8 \\
        \hline
        $\#$ of words in $\neg A$ & 11.2 & 5.3 \\
        \hline
        $\#$ of words in B & 5.4 & 1.6 \\
        \hline\hline
        \end{tabular} 
    \caption{Decriptive statistics for stories.} 
    \label{tab:descriptive} 
    \end{center} 
\end{table}

\section{Analysis of Human Beliefs about events A and B} \label{results:beliefs}
In addition to measuring the reading times that reflect online processing, we also collected the answers to the questions about occurrences of events A and B that were presented after each story (``\textit{How sure are you that event A/B happened?} -- see Figure \ref{fig:example_materials}, part III''). 

The motivation for this was to gain insights into a) how exactly subjects accommodate a causal conflict (the $\neg A \rightarrow B$ condition) and b) whether subjects indeed infer event A when it is omitted from the story (the $nil \rightarrow B$ condition). The $A \rightarrow B$ condition serves as a baseline. We analyse the collected ratings using ordinal regression models \citep{christensen2018cumulative}.



\begin{table}[t]
    \begin{center} 
    \addtolength{\tabcolsep}{-0.1em}
        \begin{tabular}{lrcrrr} 
        \hline
         & $A \rightarrow B$ & $nil \rightarrow B$ & $\neg A \rightarrow B$ \\
        \hline
        Event A & 6.41 (1.45) & 4.85 (2.89) & 3.67 (3.19)  \\
        Event B & 6.13 (1.84) & 4.91 (2.80) & 3.79 (3.13)   \\
        \hline
        \end{tabular} 
    \caption{Mean  subjects' belief ratings (and SD in parentheses) that the event actually happened in the story, by event type (A or B) and story condition ($A \rightarrow B$, $nil \rightarrow B$, and $\neg A \rightarrow B$).} 
    \label{tab:ratings} 
    \end{center} 
\end{table}

In the $A \rightarrow B$ condition, both events A and B were given on average high ratings (6.41 and 6.13, respectively -- see Table \ref{tab:ratings}), meaning that subjects were sure that the events happened when they both were explicitly mentioned in the story. Further, for both events, the ratings in the $\neg A \rightarrow B$ (\textbf{event A}: $b = -2.03$, $se = 0.24$, $z = -8.67$, $p < .001$; \textbf{event B}: $b =  -1.6$, $se = 0.2$, $z = -8.22$, $p < .001$) and $nil \rightarrow B$ (\textbf{event A}: $b = -1.46$, $se = 0.22$, $z = -6.6$, $p <.001$; \textbf{event B}: $b = -0.99$, $se = 0.2$, $z = -4.97$, $p < .001$) were significantly lower compared to the $A \rightarrow B$ condition.

The analysis of subjects’ ratings showed that the causal conflict (the $\neg A \rightarrow B$ condition) resulted in lowered beliefs about both events A and B ($3.67$ and $3.79$, respectively). One potential explanation for this is that subjects might have used different strategies to resolve the conflict. For example, some subjects could assume that event B in fact did not happen, (however, contrary to the narrative) because the premise is not met. While others could resolve the conflict by assuming that event A in fact happened thus making event B also possible to happen. Both strategies would explain relatively lower strength of beliefs about both events B and A to happen. Any explanations, however, necessitate a follow-up study with more elaborative questions that potentially require subjects to provide explanations of the given ratings. 


Interestingly, we also observe lower ratings for both events in the $nil \rightarrow B$ condition, compared to the $A \rightarrow B$ condition, which is contrary to our expectations. In the $nil \rightarrow B$ condition, event B was overtly mentioned in the story, which should lead to comparable strength in subjects' beliefs with the $A \rightarrow B$ condition. Subsequently, event A, even though not mentioned explicitly, should be inferred on the basis of the causal link between them and script knowledge: if she added star-shaped sprinkles (event B), then she should have prepared cake decorations beforehand (event A) -- see Figure \ref{fig:example_materials}, part II. 

A probable rationale for the discrepancy between our expectation and the actual ratings is that, when faced with the questions, subjects may have retrospectively re-evaluated the story, relying more on their memory representations. Compared to condition $A\rightarrow B$, event B might have been perceptually less salient in the $nil \rightarrow B$ condition. Event B is easy to integrate due to its relation to the corresponding script (which we observe in the reading time analysis -- see Section \ref{results:rt}, RQ2) and may not receive a lot of attention from the reader, hence reducing its memorization and subsequent retrieval of event $B$. In the $A \rightarrow B$ condition, on the other hand, attention to event B is strengthened by the causal link coming from an explicitly mentioned event A that might facilitate its retrieval from memory at the question answering stage \cite[see][for similar results in reading everyday stories where subjects were asked to evaluate which events were mentioned in the text]{bower1979scripts}.

\section{Details of LLMs}
\label{app:models}

\subsection{GPT models}

\myparagraph{GPT-2.} \label{GPT-2}
GPT-2 \citep{radford2019language} is one of the most influential language models by OpenAI. As a decoder-only causal PLM, GPT-2 is often used as a baseline. 

\myparagraph{GPT-3 models.} \label{GPT-3}
GPT-3 \citep{brown2020gpt3} is the upgraded version of GPT-2 which uses almost the same model and architecture but with a significantly larger amount of parameters, which was ten times more than any previous non-sparse language model. 
GPT-3 and GPT-3.5 were chosen to be evaluated as they were expected to perform the best, based on their strong performance on a range of NLP tasks. We experiment with different versions of GPT-3 and GPT-3.5.\footnote{More details are on \url{https://platform.openai.com/docs/model-index-for-researchers}} 
\textbf{GPT-3 models \citep{brown2020gpt3}:} \texttt{curie} is a GPT-3 with 6B parameters. \texttt{davinci} is a GPT-3 with 175B parameters. 
\textbf{InstructGPT models \citep{ouyang2022instructgpt}:} \texttt{davinci-instruct-beta} is a model trained with supervised fine-tuning on human demonstrations; \texttt{text-davinci-001} and \texttt{text-curie-001} further includes top-rated model samples from quality assessment by human labellers. 
\textbf{GPT 3.5 models \citep{neelakantan2022gpt3.5}:} \texttt{text-davinci-002} is an InstructGPT model based on a model trained with a blend of code and text; \texttt{text-davinci-003}  was further trained using reinforcement learning with human feedback. 

\myparagraph{Newer models from OpenAI} like GPT-4 \citep{openai2023gpt4} or \texttt{gpt-3.5-turbo-0301} don't support the "Completions" API and can't return probabilities so we don't include them. 

\subsection{Chatbots}
As the two current state-of-the-art LLMs, GPT-4 and ChatGPT, are both designed to function as chatbots, our aim is to harness the potential of the most capable open-source chatbot available to us. Chatbots, by design, need to comprehend and respond contextually to inputs, often requiring them to make connections between disparate pieces of information in a conversation.
\myparagraph{Vicuna} is an open-source chatbot created by fine-tuning an LLaMA base model with approximately 70K user-shared conversations collected from ShareGPT.com. Preliminary evaluation in their paper \citep{vicuna2023} suggests that Vicuna reaches 90\% of the quality of chatbots such as ChatGPT and Google's Bard.

\subsection{Efficient Models} \label{Efficient Models}
There are models that need less memory or less time. Methods that reduce space could have a better performance here, because, for most of this experiment, we had limited space. Efficient models are interesting for long-range dependencies because they employ innovative techniques or optimizations to handle dependencies more effectively. Efficient models might be better or worse at capturing the relationships between distant parts of the text due to their unique approaches.

Nyströmformer and language perceiver are examples of models with efficient self-attention.

\end{document}